# Time-Dependent Utility and Action Under Uncertainty

**Eric Horvitz**
Palo Alto Laboratory
Rockwell International Science Center
444 High Street
Palo Alto, California 94301

**Geoffrey Rutledge**
Medical Computer Science Group
Knowledge Systems Laboratory
Stanford University
Stanford, California 94305

## Abstract

We discuss representing and reasoning with knowledge about the time-dependent utility of an agent's actions. Time-dependent utility plays a crucial role in the interaction between computation and action under bounded resources. We present a semantics for time-dependent utility and describe the use of time-dependent information in decision contexts. We illustrate our discussion with examples of time-pressured reasoning in Protos, a system constructed to explore the ideal control of inference by reasoners with limited abilities.

## 1 INTRODUCTION

Decision-theoretic methods have been considered inapplicable for general problem solving because they require agents to possess a utility function that provides a preference ordering over outcomes of action, and to have access to a probability distribution over outcomes associated with each decision (Simon et al., 1987). We have investigated methods for maximizing utility in reasoning systems, given limitations in computational abilities and information. In particular, we have explored the problem of computing probability distributions under resource constraints. To a lesser extent, we have studied the assessment and custom-tailoring of utility models for time-dependent action.

Performing inference to determine a probability distribution can delay an agent's action. Inference-related delays can lead to losses stemming from competition for limited resources, decay of physiological states, and problems with coordination among independent decision makers. Endowing an agent with the ability to trade off the accuracy or precision of an analysis for more timely responses can increase the expected value of that agent's behavior. Growing interest and recent work by several investigators have addressed such tradeoffs in reasoning systems (Doyle, 1988; Horvitz, 1988; Boddy and Dean, 1989; Russell and Wefald, 1989; Breese and Horvitz, 1990).

We constructed the Protos system to experiment with the use of metareasoning procedures to control inference approximation methods (Horvitz et al., 1989a). Protos determines the length of time it should dwell on an inference problem before taking action in the world. Protos iteratively computes a myopic estimate of the expected value of computation (EVC) by balancing the cost of delay with the benefits expected from additional refinement of the probabilities used in a decision problem. The system makes use of information about the convergence of approximate results to exact answers, and about the time-dependent change of the utility of outcomes.

We discuss several aspects of our work on the consideration of time-dependent utility of outcomes. We review background on the Protos system, describe the semantics and assessment procedures for time-dependent utility, and discuss the custom-tailoring of default time-dependent utility models given observations. Finally, we describe the operation of Protos by presenting examples of the system's behavior.

## 2 A LIMITED REASONER

Determining the expected value of alternate actions under uncertainty requires the assignment of belief, $p(H|E,\xi)$, to one or more relevant hypotheses, $H$, given observations, $E$, and background information, $\xi$. Inference approximation algorithms produce partial results in the form of bounds or second-order probability distributions on relevant probabilities. Let us refer to relevant probabilities as $\phi$. If we are forced to act immediately, we should take an action $D$ that maximizes our expected utility, given the mean of $p(\phi)$, $<p(\phi)>$ (Howard, 1970). The utility of this action is equal to the utility of the decision we would make had belief in $\phi$ been a point probability at the mean of $p(\phi)$. That is,

$$\arg\max_D u(D, p(\phi)) = \arg\max_D u(D, <p(\phi)>)$$

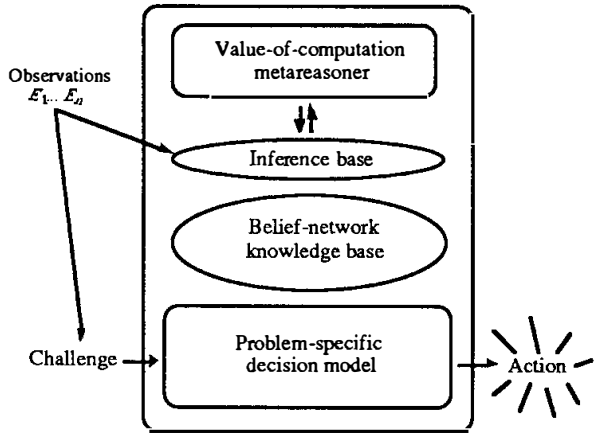

Figure 1: Protos' four components include (1) a metareasoner that considers the benefits of continuing to compute, (2) an inference base containing probabilistic inference procedures, (3) belief networks representing domain domain; and (4) a problem-specific decision model. Inference and time-dependent utility depend on observations.

Additional computation can tighten a second-order distribution. However, the utility of outcomes can diminish with time. Thus, there is a tradeoff between the benefits of making a decision based on a more precise result and the costs associated with delay. An EVC analysis compares the expected utility of instantaneous action with the expected utility of action that might be taken following future computation, including the costs of that computation.

An exact EVC analysis can consume a significant portion of the total computational cost of solving an inference problem. Our investigation on the control of belief-network inference has focused on the use of tractable EVC approximations. Approximate EVC analyses include single-step or *myopic* analyses. In myopic analyses, the EVC is computed under the assumption that an agent will take an action in the world after reasoning for a predetermined increment of time; we undertake a myopic analysis to determine if additional analysis is more valuable than immediate action. One approach to computing the expected utility delaying action is to consider the set of second-order distributions expected with additional computation. For each feasible future distribution, we consider the value of the best action, given that distribution, and weight that utility by the probability of the future distribution.

Protos makes use of myopic EVC analyses. Protos has four major components, pictured schematically in Figure 1: (1) a metareasoner; (2) an inference base containing inference procedures; (3) a domain-specific knowledge base in the form of belief networks; and (4) a problem-specific decision model. At run time, a decision problem containing alternate actions, outcomes, and utilities is passed to Protos. Given a decision problem, Protos initiates an iterative cycle of reasoning and metareasoning. Object-level inference is interleaved with metareasoning about the value of continuing to perform additional inference.

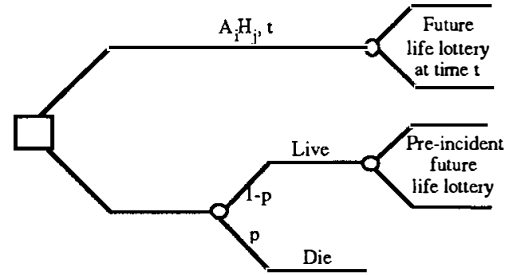

Figure 2: Lottery for assessing time-dependent utilities. We query a decision maker for the probability $p$ of instant, painless death that would make him indifferent between his future life lottery when treated at time $t$, and having a $1 - p$ chance at continuing his life as if the challenge facing him had not occurred.

At the start of each cycle, Protos computes the EVC associated with continuing object-level computation for an additional increment of time. If the metareasoner indicates that the EVC associated with the next increment of reasoning is zero or negative computation ceases and the system takes an action indicated by the mean of the second-order probability distribution. Depending on the computational hardware, the structure of the time-dependent utility model, and the expected refinement of the second-order probability distribution by an inference algorithm, Protos may (1) take an immediate reflex action, (2) dictate a best action after some partial inference, or (3) take an action it proves to be dominant. Decision dominance can be proved before inference is completed with the use of a probability bounding algorithm. A decision dominates others when a single action is indicated for the range of probabilities in the interval bordered by an upper and lower bound on the probability.

We have experimented with a tractable myopic approximation named EVC/BC (for *EVC-bounds categorical*) to control probabilistic bounding. With this form of EVC, we compute the value of tightening categorical upper and lower bounds on a probability. EVC/BC hinges on interpreting upper and lower bounds as a second-order probability distribution. The measure is based on a least-commitment interpretation of bounds as a uniform distribution between the upper and lower bounds, with a mean at the midpoint of the bounds interval. The small amount of time required for the EVC/BC analysis is included in the EVC analysis itself. Details about the nature, limitations, and use of EVC/BC are described in (Horvitz, 1990).

# 3 TIME-DEPENDENT UTILITY

Let us consider the use of Protos to solve time-pressured medical problems. We have worked to represent in Protos the cost of delaying treatment as a function of the time a patient has remained in an untreated acute pathophysiological state. Physicians delivering emergency medical care often rely on knowledge about the cost of delay in treating a patient.

## 3.1 Semantics and Representation of Time-Dependency

In answer to a query for assistance Protos propagates observations about a patient's symptomatology through a belief network. The system deliberates about whether to make a treatment recommendation immediately, based on a partial analysis, or to defer its action and to continue inference, given its knowledge about the costs of delay.

We represent time-dependent action by considering a continuum of decisions, each defined by initiating an action at a progressively later time, and by assessing the change in utility of the outcome as a function of this time. We use $A_i H_j, t$ to refer to an action, $A_i$, taken at time $t$ when state $H_j$ is true. We define $t$ in terms of an initial time, $t_o$, the time a physiological challenge begins. We define the utility of $u(A_i H_j, t)$ at different times $t$, with an *acute-challenge lottery*. To assess the cost of delaying a treatment, we ask a decision maker to consider a time-pressured problem that he might face in a decision context. Next, we imagine that there is a treatment that can rid him instantly of the acute affliction with probability $1 - p$. Unfortunately, with probability $p$, the treatment will kill him, immediately and painlessly. We assume that, if a patient wins this lottery, he will continue his life as if the acute incident had not occurred; that is, he faces his preincident future life lottery. To assess the utility, $u(A_i H_j, t)$, at progressively later times $t$ for action, we ask a decision maker for the probability $p$ of instant, painless death that would make him *indifferent* to accepting the uncertain outcome of being treated for an acute illness at time $t$ or having a $1 - p$ chance of continuing his life as if the acute incident facing him had never occurred. We take the difference in the probabilities of death for action at time $t$ and at a later time $t'$ as the loss in utility. We can measure the cost of delay in terms of micromorts. A *micromort* is a $10^{-6}$ chance of immediate, painless death. Alternatively we can assign dollar values to the risks incurred with delay. We can use the *worth-numeraire model* introduced by Howard (Howard, 1980) to convert small probabilities of death to dollars in terms of dollars per micromort.

Beyond assessing utilities for each moment of action, we can model the utility of action at progressively later times with functions that encode a micromort flux for each outcome. The micromort flux is the number of micromorts we incur with each second of delay. We experimented with parametric utility equations and found several to be useful for summarizing the time dependency of alternate outcomes. Two functions we used to model losses with time, are the linear and exponential forms,

$$u(A_i H_j, t) = u(A_i H_j, t_o)e^{-k_a t}$$

$$u(A_i H_j, t) = u(A_i H_j, t_o) - c_b t \quad \text{where } u(A_i H_j, t) \geq 0$$

where $k_a$ and $c_b$ are parameter constants derived through fitting a series of micromort assessments to a functional form or are assessed directly. Our language for assessing and representing mathematical models of time-dependence allows decision makers to encode lower bounds on utility over time, and to make statements about the chaining of sequences of functional forms.

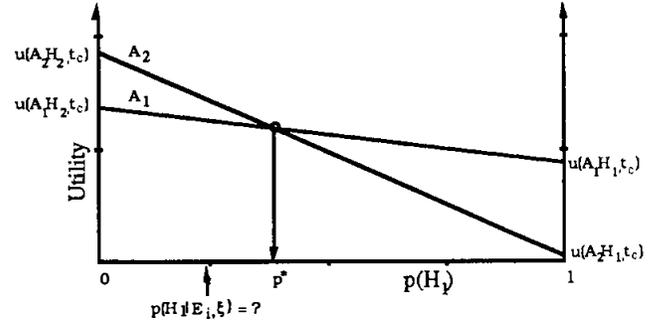

Figure 3: A graphical representation of the utility of two actions under uncertainty. The lines indicate the utilities of action $A_1$ and action $A_2$ as a function of the probability of hypothesis $H_1$. The lines cross at a threshold probability of hypothesis $H_1$ called $p^*$.

## 3.2 Utility of Action in Time-Pressured Contexts

Given time-dependent utilities, we can compute the expected value of different actions, $A_i$, in terms of the likelihood of alternative outcomes, $H_j$. The expected utility (eu) of taking action $A_i$ at time $t$ is

$$eu(A_i, t) = \sum_{j=1}^{n} p(H_j | E, \xi) u(A_i H_j, t)$$

Consider the simple case of a binary time-dependent decision problem. We have two states of the world (e.g., diseases) $H_1$ and $H_2$ and two best actions (treatments) $A_1$ and $A_2$ to address each state. As an example, the states can be the presence and absence of a disease, and the ideal actions can be treating and not treating for the disease. Under uncertainty, we must consider the utilities of four outcomes: $u(A_2 H_2, t)$, $u(A_1 H_2, t)$, $u(A_1 H_1, t)$, and $u(A_2 H_1, t)$. If $H_1$ and $H_2$

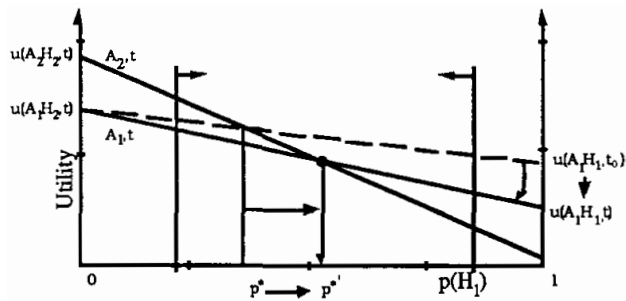

Figure 4: This graph displays how the utility of an outcome can decay as a function of time. In this case, the utility of taking action $A_1$, in the context of $H_1$, diminishes with delay. The utility associated with immediate action (broken line) and delaying action (adjacent solid line) is displayed. Note that the decision threshold, $p^*$, is also a function of time; in this case, $p^*$ increases as the utility of $A_1H_1, t$ decreases.

are mutually exclusive states, the expected utilities of the actions $eu(A_1, t)$ and $eu(A_2, t)$ are:

$$eu(A_1, t) = p(H_1|E, \xi)(u(A_1H_1, t) - u(A_1H_2, t)) + u(A_1H_2, t)$$
$$eu(A_2, t) = p(H_1|E, \xi)((u(A_2H_1, t) - u(A_2H_2, t)) + u(A_2H_2, t)$$

The expected utilities of actions $A_1$ and $A_2$, as a function of the probability of $H_1$, are graphed in Figure 3. Note that the equations specify the expected utility of two action as lines intersecting at a threshold probability of $H_1$, denoted $p^*$. As we increase the probability of $p(H_1)$ from zero to 1, the decision with the greatest expected utility shifts, at $p^*$, from $A_1$ to $A_2$. If we must act immediately, we take an action dictated by the mean of the second-order distribution: We take action $A_1$ if the mean of the second-order distribution over $p(H_1|E, \xi)$ is greater than $p^*$. Otherwise it is best to take action $A_2$.

A computational agent rarely is forced to act immediately. An agent can pause to continue inference, or to reflect about the costs and benefits of delaying an action to compute a better decision. The dynamics of reasoning about belief and action under bounded resources is highlighted in Figure 4. The figure shows how the utility of outcome $A_1H_1, t$ might diminish with delay. The dashed line shows the expected utility of taking $A_1$ in the context of $H_1$ at an initial time, $t_o$. The adjacent solid line indicates the diminished expected utility of taking the action at a later time $t$, given the truth of $H_1$. Note that, as the utility of taking action $A_1$ falls, the decision threshold, $p^*$, increases.

In a time-pressured computational setting the utility of one or more outcomes decay with delay. At the same time, inferential processes may be underway to refine bounds or a second-order distribution over probabilities of interest. Figure 4 shows the concurrent tightening of upper and lower bounds by a bounding algorithm. As the utility lines pivot or sweep down at rates dictated by the decay functions for each outcome approximate inference continues to tighten the bounds, yielding a time-dependent dynamics of belief and action.

### 3.3 Run-Time Modification of Criticality

Most of the work on Protos has relied on the use of files of utilities assessed for prototypical situations. The utility information is represented in tuples which contain the utility of immediate action, and time dependent decay, indexed by $A_iH_J$ pairs. However, we also have explored the construction of *models* of time-dependent utility. With the modeling approach, we assess utilities that represent preferences for canonical situations and apply a mathematical model to customize "average case" utilities and time-dependencies to a specific decision maker and situation. To handle time-pressured medical decisions, we elicit from an expert decision maker—in our case, an emergency-room physician[1]—functions that modify the micromort flux of relevant outcomes, in response to arguments of discrete and real-valued patient vital signs. We experimented with functions that provide time-dependency parameters as a function of the patient's age, heart rate, blood pressure, and partial pressure of oxygen in the blood (PaO2). In practice, Protos makes use of default time-dependent utility models if no vital signs are observed. Given the observation of vital signs, and the availability of information about the specific class of decision problem, the initial utility and time-dependence are customized.

Our work on customizing time-dependent utility through constructing models of criticality parallels work in the medical decision analysis community on tools for assisting physicians to induce the utility functions of patients by identifying key features of their personalities (McNeil et al., 1982; Jimison, 1990). Our experimentation with deterministic functions for modifying utility models is a modest initial approach to customizing default time-dependent models. In the general case, modeling the utility of decision makers, such as patients receiving time-critical therapy, is a problem of diagnosis under uncertainty.

---

[1]One of the authors (G.R.) has served as the source of emergency-medicine expertise.

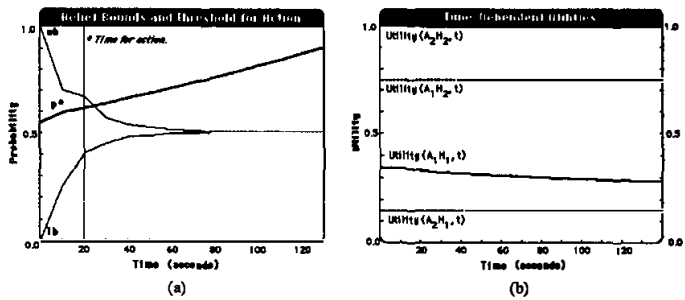
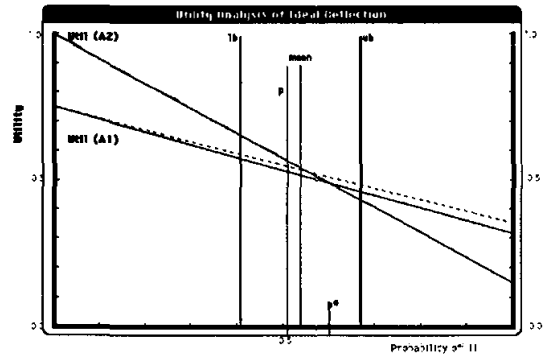

Figure 5: Time-dependent inference and ideal action. (a) Protos displays the convergence of the upper and lower bounds (ub, lb) on a probability of interest and the time-dependent decision threshold ($p^*$). The vertical line indicates the time for action. (b) The time-dependent utilities for four possible outcomes.

Figure 6: Graphical analysis of action. The utility (crossing solid lines) of treating for hypothesis $H_1$ (Util($A_1$)) and for $H_2$ (Util($A_2$)) as a function of the probability of $H_1$. Broken lines indicates the utilities of acting at $t_o$. The vertical line (p) displays the value of the exact probability, computed after the decision to take action $A_2$ was made.

## 4 PROTOS IN ACTION

We now examine the behavior of Protos in solving several simplified time-dependent decision problems in medicine. In the examples we determine the ideal time to perform inference with the bounded-conditioning approximation strategy (Horvitz et al., 1989b), given time-dependent changes in the utility of outcomes.

Bounded conditioning is based on the method of conditioning (Pearl, 1988). The method works by decomposing a belief-network inference problem into a set of simpler, singly connected belief-networks and solving these subproblems in order of their contribution to upper and lower bounds on a probability of interest. The more subproblems that are solved, the tighter the bounds. We shall examine decisions based on inference with *Dxnet* and *Alarm*, multiply connected belief networks that were assessed for reasoning about acute medical problems (Beinlich et al., 1989; Rutledge et al., 1989).[2] We note that several approximation algorithms and exact algorithms (such as the clique-tree method of Lauritzen and Spiegelhalter (Lauritzen and Spiegelhalter, 1988)) can solve inference problems with these networks faster than bounded conditioning can perform a complete analysis. However, the incremental and well-characterized convergence of bounds by bounded conditioning gives us the opportunity to explore fundamental interactions between time-dependent belief and utility, and more generally, to develop principles for optimizing the value of actions taken by an agent with limited inferential abilities. Principles of utility-directed control promise to be most valuable for controlling probabilistic inference in larger belief networks, such as the evolving QMR-DT network for internal medicine (Shwe et al., 1990).

---

[2] Alarm is a 37 node belief network; DxNet has 81 nodes.

### 4.1 Case Analyses

Figure 5(a) displays the time-dependent decision threshold, $p^*$ and the convergence of the upper and lower bounds (ub,lb) on a probability computed by bounded-conditioning with the Alarm network. Assume we are employing inference to determine the probability of a life-threatening respiratory pathophysiology ($H_1$), requiring dangerous ventilation therapy, versus a minor acute respiratory reaction that resolves in most cases with minor treatment. We assume that we shall not gather additional information; we shall base our action only on information already collected. A vertical line through the bounds in Figure 5(a) indicates Protos' decision to halt inference after 20 seconds. At this time, the EVC becomes nonpositive. Figure 5(b) displays the time-dependent utilities of four outcomes, constructed as the product of actions and states of the world: We treat ($A_1$) or do not treat ($A_2$) the patient with an invasive treatment, and the patient either has ($H_1$) or does not have the severe respiratory problem ($H_2$). The time-dependent $p^*$ is a function of the utilities, which were assessed from an expert. In this case, the utility of outcome $A_1H_1, t$— the utility of acting to treat the patient for the severe respiratory problem—decays significantly with delay.

Figure 6 displays a graph of the utility of actions $A_1$ and $A_2$ at the time action was recommended, as a function of the probability of $H_1$. The broken line, adjacent to the solid utility lines, indicates the utility of $A_1$ at $t_o$, allowing us to inspect the effect that delay has had on the value of the time-dependent outcome. The graph displays the upper and lower bounds (ub, lb) at halting time, the mean value between these bounds, and the decision threshold $p^*$ at the time Protos recommended action $A_2$. The graph also displays the final point probability of $H_1$, computed after the the entire inference problem is solved. The value of the point probability indicates that an instantaneous complete analysis would have recommended the same action. This is not always the case.

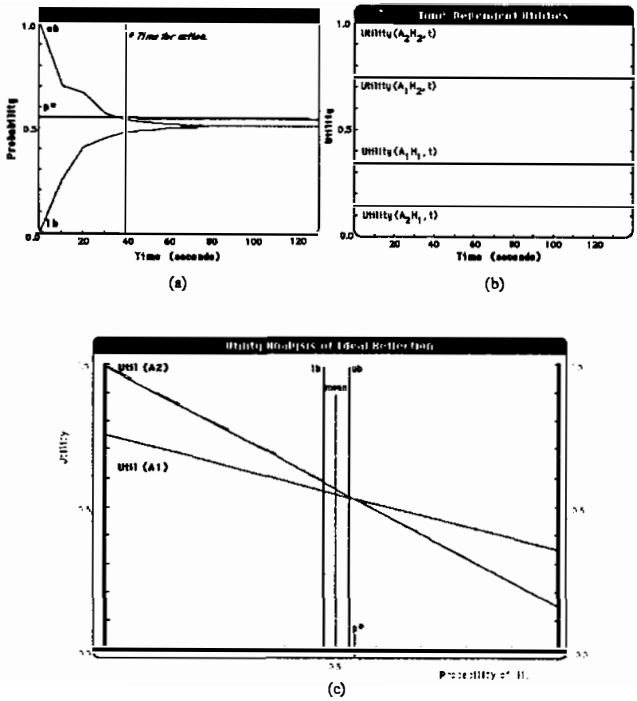

Figure 7: A less critical situation. (a) Here, decision dominance is proved as the upper bound moves below the decision threshold. (b) The time-dependent utilities for the four outcomes. (c) Graphical analysis of the bounds and utility at halting time.

To demonstrate the sensitivity of Protos' analysis to changes in time-dependent utilities, we consider the same decision problem with a smaller micromort flux for the utility of outcome, $A_1H_1,t$. Figure 7(a) displays the convergence of bounds on belief and the trajectory of the decision threshold for the revised problem. The reduced time-dependence of utilities of the outcome are displayed in Figure 7(b). With the revised utility model, which represents a less critical situation, Protos now reasons for 40 seconds before making a recommendation not to treat for $H_1$. The EVC/BC remains positive until the upper bound passes beneath $p^*$, proving the dominance of $A_2$. Figure 7(c) displays graphs of the utilities and bounds at the time action was taken.

Let us now examine Protos' performance on a cardiac decision problem with a focus on the use of default and customized utility models. Consider the case where Protos is challenged with recommending action for a patient who suddenly demonstrates extremely low blood pressure and tachycardia (an extremely fast heart rate). Assume the problem has been narrowed to two mutually exclusive syndromes: *congestive heart failure* ($H_1$) and *hypovolemia* ($H_2$). Hypovolemia is a dangerous state of decreased blood volume caused, for example, by dehydration or bleeding. Congestive heart failure (CHF) is a serious condition in which the pumping ability of the heart is weakened; like hypovolemia, it causes low blood pressure and poor oxygenation of tissues. Although hypovolemia and CHF share salient symptomatology, the treatments for these pathophysiological states conflict with each other. The treatment for hypovolemia ($A_2$) is to give the patient fluids to restore them to a normal level. In contrast, the primary treatment for CHF ($A_1$) is to reduce the quantity of liquids in the body with a diuretic. Erroneously treating a patient who has CHF with fluid-replacement therapy, or treating a patient who has hypovolemia with diuretic therapy, is life-threatening.

In Protos' default time-dependent utility model for the average case situation, the cost of delaying the treatment of CHF is described by an exponential decay constant that is ten times larger than the constant used to characterize the cost of delay in treating hypovolemia. Protos computes the probability of CHF by propagating observations in the Dxnet belief network. Figure 8(a) shows a trace of the update of the probability of CHF. Here, Protos is considering a new finding that a patient's pulmonary capillary wedge pressure is normal. (Protos was previously informed that the patient displayed low stroke volume and had low central venous pressure.) The vertical line indicates Protos' decision to halt in 115 seconds. At this point, the system recommends that the patient should be treated for CHF. The dominance of this decision is proved when the lower bound crosses the decision threshold $p^*$.

For this decision problem the micromort flux associated with delaying treatment for CHF is represented as a function of the patient's blood pressure. Let us lower the blood pressure and reevaluate the case. In response to a significant drop in blood pressure, Protos increases the exponential decay of the value for the outcome of treating for CHF, when CHF is indeed present. In this case, the decay of $u(A_1H_1,t)$ is increased from $e^{-.001t}$ to $e^{-.008t}$. Figure 8(b) shows the same probabilistic analysis with the use of the revised time-dependent utility model. Protos now recommends that the patient should be treated for CHF after only 30 seconds of computation. In the more critical case, action is indicated before a decision threshold is reached because the EVC becomes nonpositive before a probability bound crosses the decision threshold.

### 4.2 Discussion

We have made several observations about Protos' behavior. We have found that, in many cases, a utility-directed analysis of probabilistic inference dictates that actions should be taken after a small fraction of an analysis has been performed. Thus, even approximation methods with relatively slow convergence can be more valuable than faster exact algorithms. Two salient examples of this phenomena are displayed in Figure 9. In such cases the ideal decision is determined in the first few seconds of an analysis. More generally, we have found that decisions about the ideal length of time to deliberate and the ideal action to take are sensitive to the details of the time-dependent utilities of outcomes, the information about the convergence of an approximation strategy, and the trajectory of partial results generated by approximate inference.

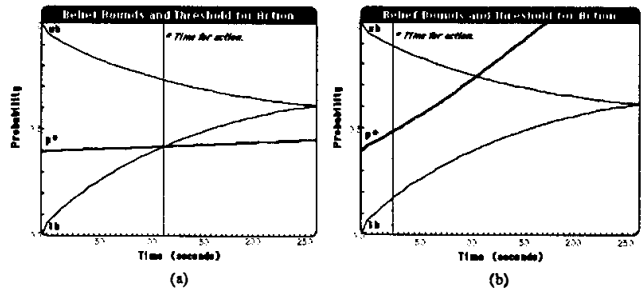
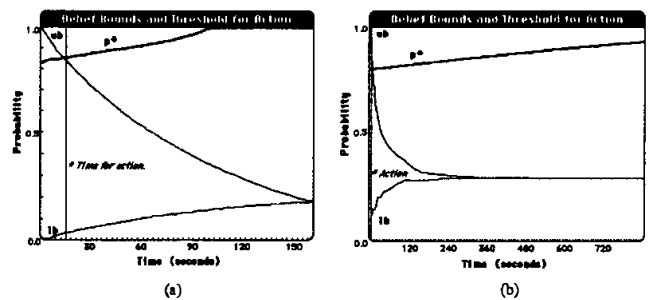

Figure 8: (a) Bounds convergence and decision threshold for decision dilemma involving treatment for CHF ($A_1$) versus for hypovolemia ($A_2$). (b) Same decision problem with increased decay of the outcome of treating for CHF when CHF is present.

Figure 9: (a) Bounds convergence and decision threshold in the Alarm network for treating possible left-ventricular failure. (b) Bounds convergence and decision threshold in reasoning within the Dxnet belief network to support a decision about treating for a pulmonary embolism.

We have observed behaviors that highlight the complexity of the interplay between time-dependent utility and time-consuming inferential processes. Some of the behaviors are explained by the limitations associated with the use of a myopic measure of EVC. We found that dependencies between time-dependent utility and inferential processes can make computation time and recommended actions sensitive to small changes in a time-dependent utility model. In some cases small changes in the time-dependencies in a utility model change the ideal recommended action.[3] We found that increasing the time-dependent decay of the utility of an outcome can *increase* the duration of reflection. In these cases the trajectory of converging bounds surrounds and "keeps step" with an increasing or decreasing $p^*$. We observed situations where an agent applying a myopic EVC estimate may be in the unlucky situation of continuing, for several steps, to observe a positive EVC, yet see its expected utility continue to diminish with delay. We identified cases where the EVC/BC returns to a positive value after it has been zero or negative. Such nonmonotonicity in the EVC motivated us to implement lookahead analyses that consider two or more future steps of computation. We are experimenting with more advanced lookahead techniques. More generally, we are pursuing the development of methods to monitor and modify behavioral patterns that have roots in the myopic EVC evaluation, and for identifying cases where the results of an analysis are sensitive to small fluctuations in the trajectory of time-dependent utilities or probabilities.

Before concluding, we stress that we have addressed the assignment of belief and utilities by limited agents; we have not discussed the automated construction of decision models. In the current version of Protos, preconstructed decision problems are passed to the system, in reaction to salient observations. We foresee that ongoing work on procedures for constructing decision models (Wellman, 1988; Breese, 1990; Heckerman and Horvitz, 1990) will foster the development of more comprehensive agents that can build as well as solve decision problems under bounded resources.

## 5 SUMMARY

We described the assessment and use of time-dependent utility in limited computational agents that are charged with taking ideal action in time-critical contexts. Analyses with Protos have demonstrated that the duration of computational analysis and the ideal decisions to make in the world can be sensitive to the time-dependent utilities of relevant outcomes. We discussed the generalization of lottery-based assessment techniques to mathematical models which represent the decay of utility of outcomes with delay. After describing the problem of customizing the time-dependency of default utility models in response to observations, we presented examples of Protos' behavior on time-pressured medical decision problems. Finally, we discussed some of Protos' behaviors and described

---

[3] Related problems with an optimal decision changing with delay for analysis have been identified previously in the context of decision analysis (McNutt and Pauker, 1987).

ongoing work on the development of nonmyopic inference monitoring and control procedures.

**Acknowledgments**

This work was supported by NASA under Grant NCC-220-51, by the NSF under Grant IRI-8703710, and by the Palo Alto Laboratory of the Rockwell International Science Center.

**References**

Beinlich, I., Suermondt, H., Chavez, R., and Cooper, G. (1989). The ALARM monitoring system: A case study with two probabilistic inference techniques for belief networks. In *Proceedings of the Second European Conference on Artificial Intelligence in Medicine, London*. Springer Verlag, Berlin.

Boddy, M. and Dean, T. (1989). Solving time-dependent planning problems. In *Proceedings of the Eleventh IJCAI*. AAAI/International Joint Conferences on Artificial Intelligence.

Breese, J. (1990). Construction of belief and decision networks. Technical Report Technical Memorandum 30, Rockwell International Science Center, Palo Alto, California.

Breese, J. and Horvitz, E. (1990). Ideal reformulation of belief networks. In *Proceedings of Sixth Conference on Uncertainty in Artificial Intelligence, Cambridge, MA*, pages 64–72.

Doyle, J. (1988). Artificial intelligence and rational self-government. Technical Report CS-88-124, Carnegie-Mellon University.

Heckerman, D. and Horvitz, E. (1990). Problem formulation as the reduction of a decision problem. In *Proceedings of Sixth Conference on Uncertainty in Artificial Intelligence, Cambridge, MA*.

Horvitz, E. (1988). Reasoning under varying and uncertain resource constraints. In *Proceedings AAAI-88 Seventh National Conference on Artificial Intelligence, Minneapolis, MN*, pages 111–116. Morgan Kaufmann, San Mateo, CA.

Horvitz, E. (1990). *Computation and Action Under Bounded Resources*. PhD thesis, Stanford University.

Horvitz, E., Cooper, G., and Heckerman, D. (1989a). Reflection and action under scarce resources: Theoretical principles and empirical study. In *Proceedings of the Eleventh IJCAI*, pages 1121–1127. International Joint Conference on Artificial Intelligence.

Horvitz, E., Suermondt, H., and Cooper, G. (1989b). Bounded conditioning: Flexible inference for decisions under scarce resources. In *Proceedings of Fifth Workshop on Uncertainty in Artificial Intelligence, Windsor, ON*, pages 182–193.

Howard, R. (1970). Decision analysis: Perspectives on inference, decision, and experimentation. *Proceedings of the IEEE*, 58:632–643.

Howard, R. (1980). On making life and death decisions. In Howard, R. and Matheson, J., editors, *Readings on the Principles and Applications of Decision Analysis*, volume II, pages 483–506. Strategic Decisions Group, Menlo Park, CA.

Jimison, H. (1990). *A Representation for Gaining Insight Into Clinical Decision Models*. PhD thesis, Stanford University.

Lauritzen, S. and Spiegelhalter, D. (1988). Local computations with probabilities on graphical structures and their application to expert systems. *J. Royal Statistical Society B*, 50:157–224.

McNeil, B. J., Pauker, S. G., Sox, H. C., and Tversky, A. (1982). On the elicitation of preferences for alternative therapies. *New England Journal of Medicine*, 306:1259–62.

McNutt, R. and Pauker, S. (1987). Competing rates of risk in a patient with subarachnoid hemorrhage and myocardial infarction: Its now or never. *Medical Decision Making*, 7(4):250–259.

Pearl, J. (1988). *Probabilistic Reasoning in Intelligent Systems: Networks of Plausible Inference*. Morgan Kaufmann, San Mateo, CA.

Russell, S. and Wefald, E. H. (1989). Principles of metareasoning. In Brachman, R. J., Levesque, H. J., and Reiter, R., editors, *Proceedings of the First International Conference on Principles of Knowledge Representation and Reasoning*, Toronto. Morgan Kaufman.

Rutledge, G., Thomsen, G., Beinlich, I., Farr, B., Kahn, M., Sheiner, L., and Fagan, L. (1989). Ventplan: An architecture for combining qualitative and quantitative computation. In *Proceedings of the Thirteenth SCAMC, Washington, DC*. IEEE Computer Society Press, Los Angeles, CA.

Shwe, M., Middleton, B., Heckerman, D., Henrion, M., Horvitz, E., Lehmann, H., and Cooper, G. (1990). A foundation for normative decision making in internal medicine: A probabilistic reformulation of QMR. Technical Report KSL-90-09, Knowledge Systems Laboratory, Stanford University.

Simon, H., Dantzig, G., Hogarth, R., Plott, C., Raiffa, H., Shelling, T., Shepsle, K., Thaler, R., Tversky, A., and Winter, S. (1987). Decision making and problem solving. *Interfaces*, 17:11–31.

Wellman, M. (1988). *Formulation of Tradeoffs in Planning Under Uncertainty*. PhD thesis, Massachusetts Institute of Technology, Cambridge, MA.